\documentclass{article}
\usepackage{spconf,amsmath,graphicx}
\usepackage{booktabs, multirow} % for borders and merged ranges
\usepackage{soul}% for underlines
\usepackage[table]{xcolor} % for cell colors
\usepackage{changepage,threeparttable} % for wide tables
\usepackage{hyperref}
\usepackage{amssymb}
\usepackage{amsmath}
\usepackage{mathtools}

\title{Analysing Discrete Self Supervised Speech \\ Representation for Spoken Language Modeling}
\name{Amitay Sicherman and Yossi Adi}
\address{School of Engineering and Computer Science\\
  The Hebrew University of Jerusalem, Israel}

\begin{document}
\maketitle

\begin{abstract}
This work profoundly analyzes discrete self-supervised speech representations (units) through the eyes of Generative Spoken Language Modeling (GSLM). Following the findings of such an analysis, we propose practical improvements to the discrete unit for the GSLM. First, we start comprehending these units by analyzing them in three axes: interpretation, visualization, and resynthesis. Our analysis finds a high correlation between the speech units to phonemes and phoneme families, while their correlation with speaker or gender is weaker. Additionally, we found redundancies in the extracted units and claim that one reason may be the units' context. Following this analysis, we propose a new, unsupervised metric to measure unit redundancies. Finally, we use this metric to develop new methods that improve the robustness of units' clustering and show significant improvement considering zero-resource speech metrics such as ABX. Code and analysis tools are available under the following \href{https://github.com/slp-rl/SLM-Discrete-Representations}{link}.
\end{abstract}


\vspace{-0.1cm}
\section{Introduction}
\label{sec:intro}
\vspace{-0.1cm}
Recently Self-Supervised Learning (SSL) methods for speech have shown great success on plenty of downs stream tasks~\cite{yang2021superb}. From Automatic Speech Recognition~\cite{hsu2021hubert, baevski2020wav2vec, riviere2020unsupervised} and speaker diarization~\cite{dissen2022self}, to phone segmentation~\cite{kreuk2020self}, these models have shown remarkable results.

Specifically, these SSL models allow recent success in Generative Spoken Language Modeling (GSLM)~\cite{lakhotia2021generative, nguyen2022generative, borsos2022audiolm}. In GSLM, we aim to learn a discrete representation of the speech signal. This is often done by applying the k-means algorithm over the continuous representation obtained from the SSL model. Then, we train a unit Language Model (uLM) over such representation, and lastly, we decode it back to a time domain signal using vocoder~\cite{polyak2021speech}. During inference time, we can sample from the uLM. 

Although these models can generate meaningful and coherent speech utterances, little is known about the properties captured but these discrete representations. The authors in~\cite{wells2022phonetic} examined the purity between phonetics elements and the discrete representations. The authors' proposed method analyzes the discrete SSL representation considering fine-grained linguistic properties, e.g., different articulatory classes or closure and release portions. The authors in~\cite{de2022probing} proposed a probing method to analyze the presence of phone classes, gender, and language information while comparing monolingual and bilingual models. 

This work analyzes quantitatively and visually discrete representations obtained by HuBERT and CPC models. Next, equipped with such an analysis, we provide a metric to identify redundancies in the discrete representations and propose a method to improve k-means clustering based on it.

We find a high correlation between the units and the phonemes, but with many redundancies in the units. We show that one reason may be the units’ context. In Addition, we propose an unsupervised metric to measure these redundancies, and we use it to significant improvement the unit clustering.
\begin{figure}[t!]
\centering
\includegraphics[width=0.8\linewidth]{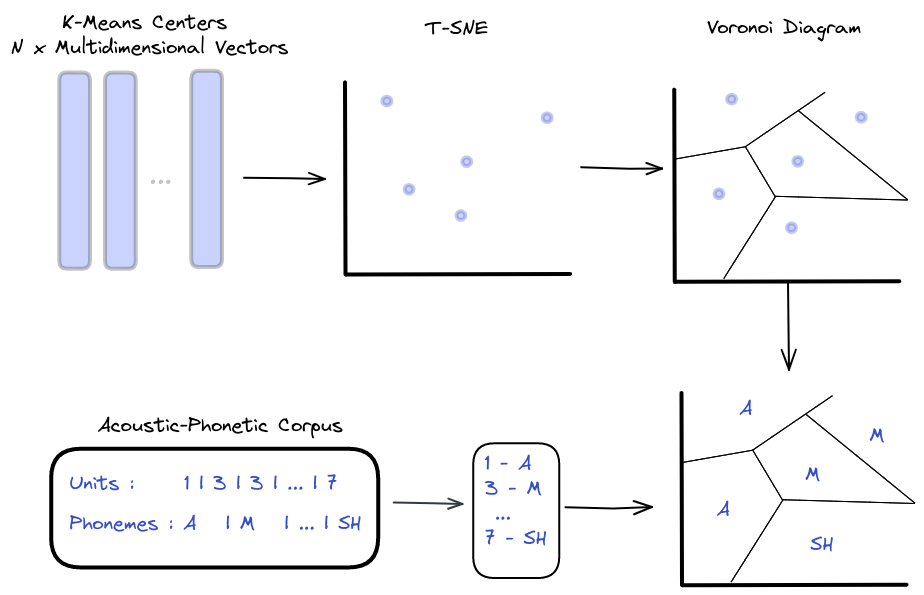}
\caption{Units visualization process. 
}
\label{fig:center-phonemes-methods}
\end{figure}

\vspace{-0.1cm}
\section{Background}
\label{sec:background}
\vspace{-0.1cm}
The general GSLM pipeline is comprised of three main modules: (i) Speech-to-unit, (ii) Unit language model, and (iii) Unit-to-speech, where each of these modules is trained separately. Speech resynthesis can be achieved while ignoring the language model and directly feeding the quantized units into the unit-to-speech module~\cite{polyak2021speech}

\begin{figure*}[t!]
\centering
\includegraphics[width=0.9\linewidth]{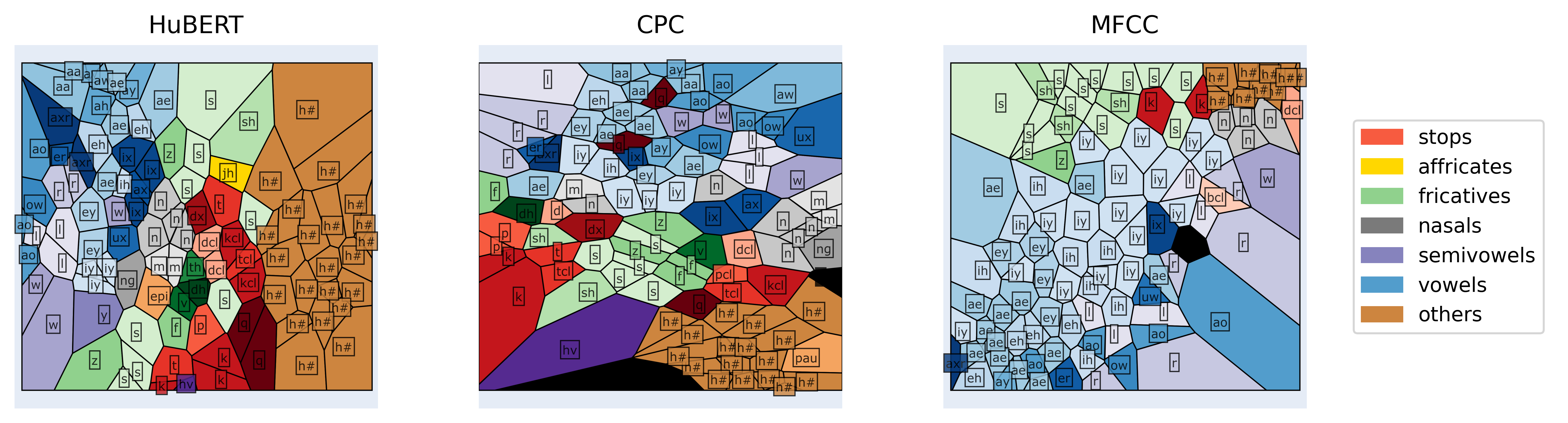}
\caption{Units visualization. Each bounded area represents a single unit and is colored by the unit's phoneme/phoneme family.}
\label{fig:center-phonemes}
\end{figure*}

\noindent {\bf Speech To Unit (STU)} The model first encodes the raw speech signal into a continuous representation and then quantizes the representation to a sequence of discrete units ~\cite{lakhotia2021generative, kharitonov2022textless, gat2022robustness}.

Formally, denote the domain of audio samples by $x \subset R$. Therefore, the representation for a raw signal is a sequence of samples $x = (x_1,\ldots, x_T)$, where  $x_t\in x$ for all $1\leq t \leq T$. Consider an encoder network, $f$, that gets as input the speech utterance and outputs a sequence of spectral representations sampled at a low frequency as follows $f(x) = (v_1, \dots, v_{T'})$. Note that we do not assume anything about the structure of the encoder network $f$. 
Since the representations learned by such models are usually continuous, a k-means algorithm is applied over the models' outputs to generate discrete units, denoted as $z = (z_1,\ldots,z_{T'})$. Each element $z_i$ in $z$ is a positive integer, $z_i\in\{1,..,K\}$ for $1\le i \le T'$, where $K$ is the number of discrete units. 

As the quantized representation, $z$, usually contain units repetitions that degrade the performance of the language modeling, a common approach is to collapse repetitions and generate a de-duplicated sequence while additionally storing the units' duration separately. For instance, the sequence \texttt{12,12,25,31,31,31} will be converted into \texttt{12,25,31} and the corresponding durations \texttt{2,1,3}.

\noindent {\bf Unit Language Model (ULM) }  is trained on the extracted and deduplicated discrete units, $z$. The language model can be used, for example, to generate speech conditionally or unconditionally.

\noindent {\bf Unit To Speech} module converts the discrete speech representation, $z$, to a raw waveform. The authors in~\cite{lakhotia2021generative} used a Tacotron2.0~\cite{shen2018natural} based model followed by WaveGlow~\cite{prenger2019waveglow} vocoder. Later, \cite{polyak2021speech} proposed a unit-based vocoder based on the HiFi-GAN architecture to convert units to speech directly. In this work, we focus on the latter setting. 

\vspace{-0.1cm}
\section{Method}
\vspace{-0.1cm}
We analyze representations obtained by either HuBERT~\cite{hsu2021hubert} or CPC~\cite{riviere2020unsupervised} models considering a various number of clusters.

\vspace{-0.1cm}
\subsection{Analysis}

\label{sec:units-phonemes}
\vspace{-0.1cm}

\noindent {\bf Units Interpretation.} 
We start by measuring the mutual information between the discrete representation and different speech properties (phonemes, speaker id, and gender) using the V-Measure score~\cite{rosenberg2007v}. The v-Measure score is on a scale from 0 to 100, where 100 represents a perfect match. The v-measure can be considered a normalized mutual information score.

For this purpose, we align each utterance with its corresponding attribute. To get units-to-phonemes alignment, we iterate over the TIMIT corpus \cite{garofolo1993timit} (which has phoneme-to-signal alignment) and count all the units-to-phonemes pairs. Next, we assign a representative phoneme to each unit based on the majority in the unit - i.e., the phoneme label is the phoneme that appears with the unit the most. For speaker and gender analysis, we use the LibriSpeech corpus as it contains a large and diverse set of speakers. 

\noindent {\bf Units Visualization.}
An additional point of view of the units' meaning is the spatial structure. For this purpose, we create a 2D spatial view containing information regarding the relationship between continuous representation, discrete units, and corresponding phonemes. Specifically, we apply the following two steps: (i) We project the high-dimensional speech representation into 2D using the T-SNE~\cite{JMLR:v9:vandermaaten08a} with the perplexity of 30.0 and 1000 iterations. T-SNE is a nonlinear dimensionality reduction that intuitively preserves the nonlinear distance relations between neighbors in the high and low dimensions. Then, we use the Voronoi diagram~\cite{aurenhammer1991voronoi} that converts the scatter plot into an area plot. Finally, we have left with a bounded area in the 2D space for each unit; (ii) In the second part, we create a single label to represent each cluster. First, we use the units-phonemes alignment from the TIMIT (similar to the process in the previous paragraph). Then, we assign for each cluster the most represented phoneme in it. Finally, we replace the unit id with their corresponding phonemes and color the area base on the phoneme and phoneme family. A visual description of the proposed method can be seen in Figure~\ref{fig:center-phonemes-methods}.

\noindent {\bf Units Resynthesis.} Next, we analyze the units' information from the opposite direction - that is, through speech resynthesis. We decode the units back to speech using a look-up-table of the corresponding 20ms speech segments, then we transcribe the generated audio and measure the transcription error. Intuitively, in case of a strong correlation between the units and the phonemes - we can take a single 
"sound'' to represent each unit -  and apply the UTS step using the concatenation of these sound pieces. Notice that this approach differs from the one in~\cite{polyak2021speech} as there is no neural vocoder. 

Formally, let $u, l$ be sequences of deduplicated units and their length obtained by applying STU on the input audio $x$. and let $x_i$ be part in $x$ that is matched to the deduped unit, $u_i$. Notice $x_i$ can be of arbitrary length. 
\\
Lookup Vocoder defines as :
\begin{equation}
    \label{eq:lv}
    \begin{aligned}
        &LV(u,l)=\texttt{concat}(F(u_1,l_1),\ldots,F(u_n,l_n)),\\
        &F(u_i,l_i)=
            \begin{cases}
                T[Key(u_i,l_i)],& \text{if } Key(u_i,l_i)\text{ in T}\\
                x_i , & \text{else}\\
        \end{cases},
   \end{aligned}
\end{equation}
where\textit{} $T$ is a Look-up-table that stores for each key the corresponding $x_i$ of the first appearance of this key, and $Key$ maps unit and length into key. Note that the lookup table is filled along a random order iteration over the dataset.

We also measure the "memorization" of the process as the mean percentage of unseen keys in the lookup table.

We explore four different types of $Key$ : (i) \textbf{Local-Single}- $Key(u_i)=(u_i$); (ii) \textbf{Local-Full}- $Key(u_i)=(u_i,l_i$); (iii) \textbf{Context-Single}- $Key(u_i)=(u_{i-1},u_i,u_{i+1}$); (iv) \textbf{Context-Full}- $Key(u_i)=(u_{i-1},u_i,u_{i+1},l_i$).

We analyze representations obtained by either HuBERT~\cite{hsu2021hubert} or CPC~\cite{riviere2020unsupervised} models considering a various number of clusters.

\begin{table}[t!]
\centering
\caption{Units Interpretation results. For phonemes, higher is better. While for the speaker and gender, a lower score indicates that the units success 'ignores' this information.}
\label{tab:v-measure}
\resizebox{.8\linewidth}{!}{
\begin{tabular}{c|c|cc|c}\toprule
\textbf{Model} &\textbf{Size} &\textbf{Speaker} &\textbf{Gender} &\textbf{Phoneme} \\\midrule
\multirow{4}{*}{\ul{CPC}} &50 &1.35 &0.66 &47.30 \\
&100 &2.35 &0.54 &\textbf{48.45} \\
&200 &3.70 &1.62 &47.74 \\
&2000 &10.39 &4.14 &44.06 \\\midrule
\multirow{4}{*}{\ul{HuBERT}} &50 &0.73 &0.03 &42.49 \\
&100 &1.41 &0.17 &45.48 \\
&200 &1.95 &0.21 &46.64 \\
&2000 &5.15 &0.65 &43.32 \\\midrule
\multirow{3}{*}{\ul{MFCC}} &50 &9.11 &2.90 &8.57 \\
&100 &11.54 &3.97 &8.73 \\
&200 &13.81 &4.59 &8.96 \\
\bottomrule
\end{tabular}}
\vspace{-0.3cm}
\end{table}
\begin{table}[t!]
\centering
\caption{Units resynthesis character error rate (CER) and mean percentage of unseen keys (Memorization). The table contains results for different lookup key types: Local-Single (\textbf{L-S}), Local-Full (\textbf{L-F}), Context-Single \textbf{(C-S)} and Context-Full (\textbf{C-F}).}
\label{tab:lookup-table}
\resizebox{.9\linewidth}{!}{
\begin{tabular}{c|c|c|cccc}\toprule
\multicolumn{7}{c}{\textbf{CER}} \\\midrule
\textbf{Model} &\textbf{Size} &\textbf{Hifi-GEN} &\textbf{C-F} &\textbf{C-S} &\textbf{L-F} &\textbf{L-S} \\\cmidrule{1-7}
\multirow{3}{*}{\ul{CPC}} &50 &5.95 &9.12 &25.36 &39.57 &60.98 \\
&100 &5.67 &6.52 &15.21 &22.51 &53.59 \\
&200 &5.37 &5.12 &10.16 &15.18 &40.65 \\\midrule
\multirow{3}{*}{\ul{HuBERT}} &50 &7.31 &10.31 &14.96 &47.24 &58.42 \\
&100 &4.39 &5.24 &6.26 &26.55 &57.49 \\
&200 &4.10 &4.25 &4.69 &15.56 &19.88 \\\midrule
\multirow{3}{*}{\ul{MFCC}} &50 &50.47 &33.85 &57.60 &71.43 &69.22 \\
&100 &44.68 &15.79 &46.55 &67.54 &66.13 \\
&200 &41.67 &6.22 &30.47 &61.46 &61.31 
\\\midrule \multicolumn{7}{c}{\textbf{Memorization}}  \\\midrule
\textbf{Model} &\textbf{Size} &\textbf{Hifi-GEN} &\textbf{C-F} &\textbf{C-S} &\textbf{L-F} &\textbf{L-S} \\\cmidrule{1-7}
\multirow{3}{*}{\ul{CPC}} &50 &- &2.66 &0.17 &0.043 &0.006 \\
&100 &- &3.74 &0.35 &0.062 &0.007 \\
&200 &- &5.52 &0.72 &0.085 &0.010 \\\midrule
\multirow{3}{*}{\ul{HuBERT}} &50 &- &2.35 &0.33 &0.028 &0.006 \\
&100 &- &3.46 &0.75 &0.039 &0.008 \\
&200 &- &4.91 &1.49 &0.056 &0.012 \\\midrule
\multirow{3}{*}{\ul{MFCC}} &50 &- &3.45 &0.29 &0.038 &0.005 \\
&100 &- &7.09 &1.01 &0.052 &0.006 \\
&200 &- &12.56 &2.50 &0.073 &0.009 \\
\bottomrule
\end{tabular}
}
\vspace{-0.3cm}
\end{table}
 \begin{table*}[t!] \centering
\caption{Comparing the different clustering methods using  ABX and speaker information. For both metrics, lower is better.}
\label{tab:improve-table}
\resizebox{.9\linewidth}{!}{
\begin{tabular}{c|c|cccc|cccc|cccc}\toprule
\multirow{2}{*}{\textbf{Model}} &\multirow{2}{*}{\textbf{Size}} &\multicolumn{4}{c|}{\textbf{ABX within}} &\multicolumn{4}{c|}{\textbf{ABX across}} &\multicolumn{4}{c}{\textbf{Speaker probing}} \\\cmidrule{3-14}
& &k-means &K-K &K-H &K-WH &k-means &K-K &K-H &K-WH &k-means &K-K &K-H &K-WH \\\midrule
\multirow{3}{*}{\ul{CPC}} &50 &5.66 &\textbf{5.38} &9.62 &8.80 &7.83 &\textbf{6.77} &11.46 &10.56 &42.22 &32.96 &19.26 &\textbf{18.15} \\
&100 &\textbf{5.42} &5.44 &6.66 &6.04 &\textbf{7.07} &7.13 &8.26 &7.49 &52.96 &45.19 &20.37 &\textbf{15.56} \\
&200 &5.53 &\textbf{5.27} &5.61 &5.68 &7.35 &\textbf{7.10} &7.28 &7.13 &63.70 &49.63 &26.30 &\textbf{22.59} \\\midrule
\multirow{3}{*}{\ul{HuBERT}} &50 &7.23 &5.67 &\textbf{5.94} &6.12 &8.93 &\textbf{6.83} &7.43 &7.67 &\textbf{30.37} &36.30 &36.67 &31.85 \\
&100 &5.82 &\textbf{5.01} &5.30 &5.29 &7.47 &6.50 &6.54 &\textbf{6.32} &48.15 &48.89 &48.15 &\textbf{46.67} \\
&200 &5.79 &5.24 &5.18 &\textbf{5.05} &7.49 &6.42 &6.46 &\textbf{6.07} &65.19 &61.11 &\textbf{54.81} &62.96 \\
\bottomrule
\end{tabular}}
\vspace{-0.3cm}
\end{table*}
 
\subsection{Circular Resynthesis}
We introduce the Circular Resynthesis (CR) method, an utterly unsupervised evaluation metric that aims to measure the redundancies in the discrete units. 
We first perform a complete resynthesis procedure, in which we encode and decode the speech signal. Then, we apply an 
additional STU stage and measure the Unit-Edit-Distance (UED) between the first and the second units representing the speech. This metric was recently proposed by~\cite{gat2022robustness} to evaluate the robustness of discrete speech representation against signal variations. Intuitively, a high UED indicates redundancies in the discrete units. To reach the final CR metric, for each pair of units, we calculate the percentage of swaps between them over all the dataset's transcriptions.

\subsection{Robust Clustering}
Equipped with the CR metric, we explore three simple methods to improve the k-means clustering quality. In all three methods, we start from the standard k-means with $k=2000$ and iteratively merge the clusters to reach the target number of clusters. The first method, named {\bf Double K-means (K-K)}. In which we apply an additional k-means over the cluster centroids from the first k-means step. In the second method, denoted as {\bf K-means with Hierarchical Clustering (K-H)}, we apply an agglomerative clustering over the cluster centroids from the first k-means step. The last method, named {\bf K-means with Weighed Hierarchical Clustering (K-WH)}, we use an agglomerative clustering using a modified version of the euclidean distance, weighted by the CR metric. Formally, the distance metric is defined as follows: 
\begin{equation}
    \label{eq:swap}
        D(i,j)  = L2(c_i,c_j)\cdot[1-\frac{CR(u_i,u_j) + CR(u_j, u_i)}{2}]
\end{equation}
while $c_i,c_j$ are the i$^{th}$ and j$^{th}$ cluster continuous centroids, and $u_i,u_j$ are the i$^{th}$ and j$^{th}$ discrete unit. 

The intuition behind the last method is that clusters with high CR are more likely to be merged. Therefore, based on the understanding that CR measures redundancies (or similar acoustic concepts), these methods reduce redundancies and combine clustering representing similar acoustic concepts.

\vspace{-0.1cm}
\section{Results}
\vspace{-0.2cm}
\label{sec:res}
\subsection{Datasets}
We use the Librispeech\cite{panayotov2015librispeech} corpus to learn the k-means clustering (\texttt{train-clean-100}), and the \texttt{test-clean} to evaluate both the clustering methods and the lookup vocoder. We also use the Librispeech corpus to calculate the V-Measure for speaker and gender. For computing the V-Measure over phonemes, we use the TIMIT benchmark. 

\subsection{Analysis}
\label{sec:res-analysis}
\vspace{-0.1cm}
\noindent{\bf Units Interpretation.}
\label{sec:res-v-measure}
Table~\ref{tab:v-measure} presents the V-Measure results regarding three different attributes - speaker, gender, and phoneme. The V-Measure for the speaker and gender scores is lower than the score of the phonemes, indicating a high correlation to the phonemes and a low correlation to the speaker or gender. In addition, when we check the effect of the number of the units- while for the speaker/gender, more units lead to a higher score, in the phoneme score, there is a max point both for the HuBERT and CPC configurations. Therefore, we claim that redundancies cause this trend in the units. Finally, we can see that CPC has a higher score for the phonemes- but also a higher score for speaker and gender.

\noindent{\bf Units Visualization.}
\label{sec:res-spatial}
Figure~\ref{fig:center-phonemes} shows the spatial structure of the units. One can see that there is a very consistent structure- first, units that represent the same phoneme are usually close to each other. Moreover, phonemes from the same family (affricates, fricatives, Etc.' ) tend also to be close to each other. In addition, we can see that while for HuBERT and CPC, the space divide between the different phonemes families is generally equal, in the MFCC model, almost all the space uses for vowels. Notice redundancies in the clusters can also be observed from such figures. 

\noindent{\bf Units Resynthesis.}
\label{sec:res-lhv}
In Table \ref{tab:lookup-table}, we show the results for the units' resynthesis. We can see that for some configurations, there is a slight difference between the HiFi-GAN and the lookup scores- this strengthens our understanding that units express fixed sounds and are mainly correlative to phonemes. We can see that the context of the units critically affects the results, while the unit's length has a lower effect. Finally, this understanding may help understand units' redundancies, i.e., the same phoneme in a different context will represent different units. In addition, we can see a correlation between CER and memorization. However, it is essential to note that the model architecture and the lookup method also significantly impact the overall performance.

\vspace{-0.2cm}
\subsection{Robust Clustering}
\vspace{-0.2cm}

\label{sec:res-clustering}
We evaluate the proposed approach in two axes: \textbf{ABX}- We use the discrete representation and the cosine similarity metric to measure ABX within and across~\cite{kahn2020libri}; \textbf{speaker information}- We measure the accuracy of a simple DNN to predict the speaker given the discrete representation, similarly to~\cite{kharitonov2022textless}\footnote{The model architecture is an encoding layer (dim 32) followed by a transformer (4 heads, hidden size 128, dropout 0.1) and a linear layer. Librispeech `dev-clean' was used for training and `test-clean' for evaluation. We applied STU on each utterance and used the units as input. We trained the model for 100 epochs using NLL loss and Adam optimizer (lr=0.001).}. 

Table~\ref{tab:improve-table} summarizes the results. We can see that the proposed methods improve both metrics outcomes for most of the configurations. Furthermore, the best results for ABX-across were obtained using CR- this strengthens our claim regarding the unit's redundancies.

\vspace{-0.2cm}
\section{Conclusion}
\vspace{-0.2cm}
\label{sec:conc}
In this work, we analyzed the GSLM discrete unit from three different and complementary points of view: interpretation, visualization, and resynthesis. The analysis showed a strong correlation between the units and the phonemes. In addition, we found redundancies in the units, which the units' context can explain. Finally, based on these understandings, we proposed methods that improve the unit's clustering.

\section*{Acknowledgements}
We would like to acknowledge support for this research from the Israeli Science Foundation (ISF grant 2049/22).

\clearpage
\bibliographystyle{IEEEbib}
\bibliography{refs}
\end{document}